\documentclass[10pt,twocolumn,letterpaper]{article}

\usepackage{cvpr}
\usepackage{graphicx}
\usepackage{amsmath}
\usepackage{amssymb}
\usepackage{booktabs}

\usepackage{color}

\usepackage{epstopdf}
\usepackage{multirow}
\usepackage{amsmath,amssymb}
\usepackage{algorithm}
\usepackage{algorithmicx}
\usepackage{algpseudocode}

\usepackage{booktabs}
\usepackage{makecell}
\usepackage{textcomp}
\usepackage{mathrsfs}
\usepackage{multicol}
\usepackage{mdframed}

\usepackage[pagebackref,breaklinks,colorlinks]{hyperref}

\usepackage[capitalize]{cleveref}
\crefname{section}{Sec.}{Secs.}
\Crefname{section}{Section}{Sections}
\Crefname{table}{Table}{Tables}
\crefname{table}{Tab.}{Tabs.}

\begin{document}

%%%%%%%%% TITLE - PLEASE UPDATE
\title{JoyGen: Audio-Driven 3D Depth-Aware Talking-Face Video Editing}

\author{
    Qili Wang\textsuperscript{1}, Dajiang Wu\textsuperscript{1}, Zihang Xu\textsuperscript{2}, Junshi Huang\textsuperscript{1}, Jun Lv\textsuperscript{1} \\
    \textsuperscript{1}JD.Com, Inc., \texttt{\{wangqili, wudajiang, huangjunshi1, lvjun\}@jd.com} \\
    \textsuperscript{2}The University of Hong Kong, \texttt{xuzihang@connect.hku.hk} \\
    \tt \url{https://joy-mm.github.io/JoyGen}
}

\maketitle

\begin{figure*}       %不带*单栏，带*双栏
    \centering
    \includegraphics[scale=0.45]{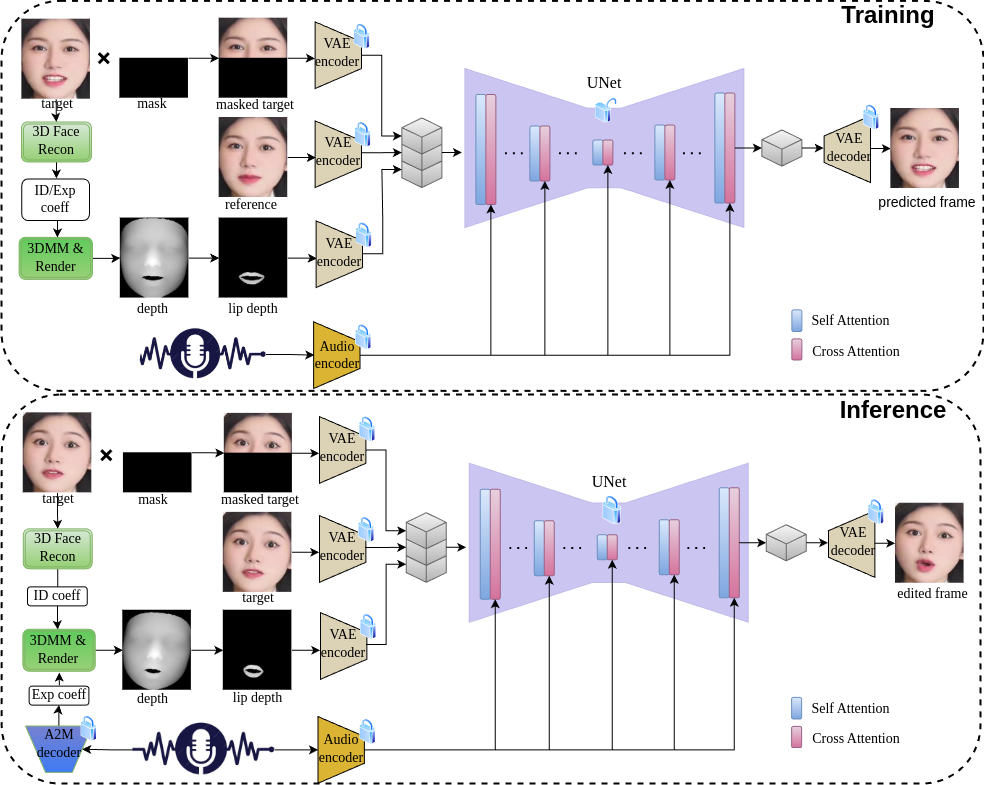}  %scale缩放比例，Fig.jpg文件名
    \caption{Overview of the proposed method JoyGen. \textbf{Top}: Training pipeline, \textbf{Bottom}: Inference pipeline.}   % 图片名称
    \label{fig:framework}
\end{figure*}

%%%%%%%%% ABSTRACT
\begin{abstract}
Significant progress has been made in talking-face video generation research; however, precise lip-audio synchronization and high visual quality remain challenging in editing lip shapes based on input audio. This paper introduces JoyGen, a novel two-stage framework for talking-face generation, comprising audio-driven lip motion generation and visual appearance synthesis. In the first stage, a 3D reconstruction model and an audio2motion model predict identity and expression coefficients respectively. Next, by integrating audio features with a facial depth map, we provide comprehensive supervision for precise lip-audio synchronization in facial generation. Additionally, we constructed a Chinese talking-face dataset containing 130 hours of high-quality video. JoyGen is trained on the open-source HDTF dataset and our curated dataset. Experimental results demonstrate superior lip-audio synchronization and visual quality achieved by our method. The training and inference code, model weights, and sample videos are available at \url{https://github.com/JOY-MM/JoyGen}.
\end{abstract}

%%%%%%%%% BODY TEXT
\section{Introduction} \label{sec:intro}
In recent years, diffusion models have demonstrated strong generative capabilities in the fields of image, video and audio generation. In the task of generating videos from a single portrait image, diffusion models combined with temporal motion modules have achieved impressive results, producing high-quality videos that closely resemble the reference subject's appearance \cite{tian2024emo,wei2024aniportrait,chen2024echomimic,xu2024hallo,jiang2024loopy}. The generated videos exhibit natural head movements, rich facial expressions, and strong lip synchronization with audio. However, when working with existing talking videos where head poses and eye movements are already realistic and expressive, there is no need to generate these aspects. Instead, the task shifts to focusing solely on the editing and refinement of the lip movements. Current approaches that integrate diffusion models with temporal motion modules encounter challenges. Without the temporal module, videos generated from a single image exhibit noticeable frame discontinuities and poor lip-audio synchronization. Conversely, including the temporal module results in head and lip movements that are randomly sampled from a learned motion distribution conditioned on the audio signal, which does not correspond to the actual motions in the video.

To address the challenges of precise lip-audio synchronization and high visual quality in talking-face video editing, we propose JoyGen, a novel two-stage framework. In the first stage, we utilize a 3D reconstruction model to predict identity coefficients and an audio2motion model to infer expression coefficients, enabling accurate lip motion generation. In the second stage, we integrate audio features with a facial depth map to provide comprehensive supervision for generating precise and synchronized lip movements with high-quality visual appearance.

The main contributions of this paper include: (1) a novel two-stage framework for talking-face generation that integrates audio-driven lip motion and visual synthesis; (2) a method for combining audio features with a facial depth map for precise lip-audio synchronization; (3) the construction of a high-quality Chinese talking-face dataset with 130 hours of videos; and (4) state-of-the-art results in lip-audio synchronization and visual quality demonstrated on both the open-source HDTF dataset and our curated dataset.

\section{Related Works}
\textbf{Portrait Animation.}
Talking-face video generation can be broadly categorized into two categories: generating videos from a single image and editing the lip movements in existing videos. Techniques like EMO\cite{tian2024emo}, Aniportrait\cite{wei2024aniportrait}, EchoMimic\cite{chen2024echomimic}, Hallo\cite{xu2024hallo} and Loopy\cite{jiang2024loopy} utilize Stable Diffusion (SD) as the foundational framework, incorporating networks such as ReferenceNet\cite{hu2023animateanyone} to produce video frames that closely resemble the appearance of a reference portrait. Motion smoothness and content consistency in the generated video are achieved through the temporal module reported in AnimateDiff\cite{guo2023animatediff}, which leverages a temporal Transformer architecture. By modeling temporal dependencies between features at the same spatial location along the temporal axis, it ensures the smooth transition between adjacent video frames. As a key aspect of facial dynamics, lip movements are effectively captured.

\textbf{Facial Video Editing.} Wav2Lip\cite{Prajwal_Mukhopadhyay_Namboodiri_Jawahar_2020}, VideoReTalking\cite{Cheng_Cun_Zhang_Xia_Yin_Zhu_Wang_Wang_Wang_2022}, SyncTalkFace\cite{park2022synctalkfacetalkingfacegeneration} and MuseTalk\cite{musetalk}  use a binary mask to occlude the mouth region of target frames, while selecting one or more frames to provide identity-related facial information for reconstructing the occluded areas. HyperLips\cite{Chen_Yao_Li_Wang_Zhang_Yang_Wen} comprises two phases: initially, facial frames are generated based on audio signals; subsequently, facial key points extracted from these frames are employed to enhance the quality of the final facial reconstruction. Similarly, \cite{Zhong_Fang_Cai_Wei_Zhao_Lin_Li,zhong2024stylepreservinglipsyncaudioaware,zhong2024stylepreservinglipsyncaudioaware} also consist of two phases. In the first phase, intermediate representations such as key points, 3D facial meshes, or 3D expression coefficients are used to predict lip movements from audio signals.  These representations are then integrated as additional inputs for subsequent facial generation. GeneFace++\cite{Ye_He_Jiang_Huang_Huang_Liu_Ren_Yin_Ma_Zhao_2023} and Real3DPortrait\cite{ye2024realdportrait} leverage a VAE model to map audio features to motion by predicting the expression coefficients of a 3DMM\cite{Blanz_Vetter_1999,Paysan2009A3F}. These expression coefficients are then used to construct the 3DMM face mesh, which is rasterized to obtain the PNCC\cite{Zhu_2019}. Finally, the PNCC serves as a rendering condition for generating the final face.

\section{Dataset}
Most publicly available Talking Face datasets focus predominantly on English-speaking scenarios. To promote applications in Chinese-speaking environments, we have constructed a high-definition Talking Face dataset featuring Chinese-language videos. To ensure the dataset's quality, we implemented a rigorous manual curation and screening process. Videos were sourced from the Bilibili and Douyin platforms, with playback quality maximized to the highest available resolution. The selection criteria included: (1) only one video per account to ensure diversity, (2) a single visible face per video, (3) alignment of audio with the speaker's identity, (4) clear visibility of the mouth region or teeth, and (5) Chinese audio without significant background music or noise interference.

\begin{figure*}
    \centering
    \begin{subfigure}{0.24\textwidth}   % 设置子图的宽度
        \centering
        \includegraphics[width=\linewidth]{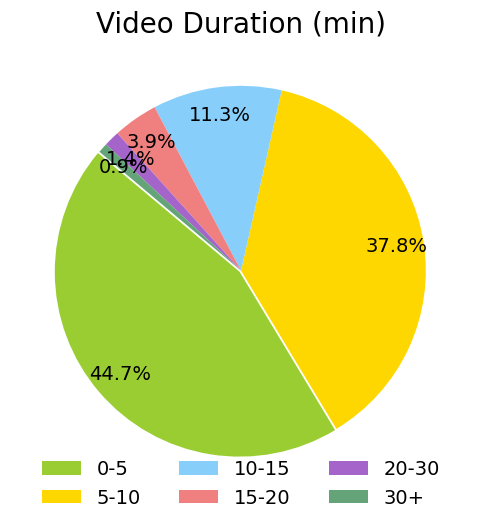}  % 替换为图片路径
        \label{fig:dataset_duration}
    \end{subfigure}
    \begin{subfigure}{0.24\textwidth}
        \centering
        \includegraphics[width=\linewidth]{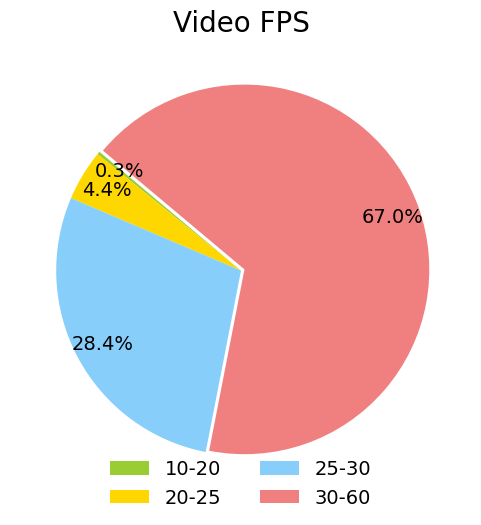}  % 替换为图片路径
        \label{fig:dataset_fps}
    \end{subfigure}
    \begin{subfigure}{0.24\textwidth}
        \centering
        \includegraphics[width=\linewidth]{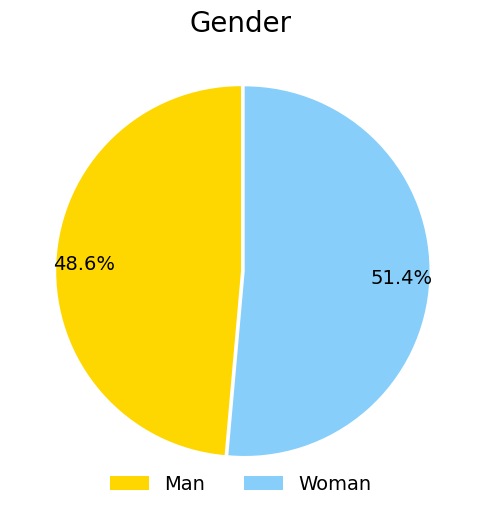}  % 替换为图片路径
        \label{fig:dataset_fps}
    \end{subfigure}
    \begin{subfigure}{0.24\textwidth}
        \centering
        \includegraphics[width=\linewidth]{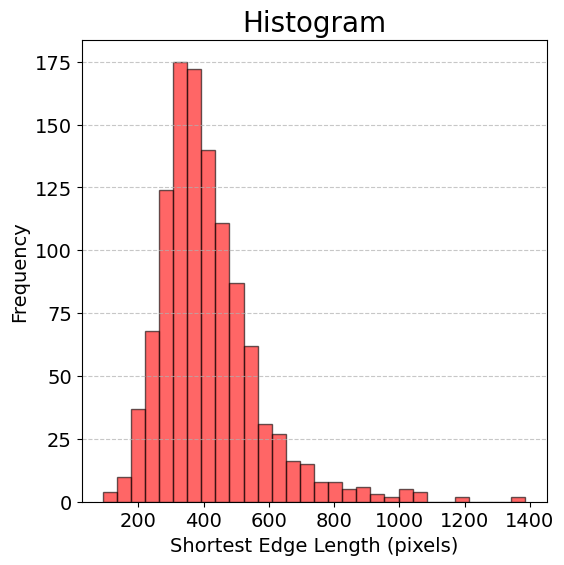}  % 替换为图片路径
        \label{fig:dataset_fps}
    \end{subfigure}
    \caption{Statistics of our newly curated chinese talking-face dataset. Our dataset has an approximately equal ratio of males and females, with varying video lengths and frame rates, and includes high-resolution video frames.}
    \label{fig:dataset}
\end{figure*}

The resulting dataset comprises approximately 1.1k videos, with durations ranging from 46 seconds to 52 minutes, and a total length of approximately 130 hours. A comprehensive statistical analysis was conducted on the dataset, as shown in Fig. \ref{fig:dataset}, covering video durations, frame rates, gender distribution and face sizes.

\section{Method}
In this section, we introduce the framework for talking-face generation, as shown in Fig. \ref{fig:framework}.
\subsection{3DMM}
\cite{Blanz_Vetter_2003} propose the 3D morphable model (3DMM) which describes a 3D face space with PCA:
\begin{equation}
\mathcal{S} = \bar{\mathcal{S}} + \alpha \mathcal{U}_{id} + \beta \mathcal{U}_{exp}
\label{eq:3dmm}
\end{equation}
where $\bar{\mathcal{S}} \in \mathbb{R}^{3N}$ is the average shape of 3D face, $\mathcal{U}_{id}$ and $\mathcal{U}_{exp}$ are the orthogonal basis of identity and expression. The coefficients $\alpha \in \mathbb{R}^{80}$ and $\beta \in \mathbb{R}^{64}$ control facial identity and expression, respectively. We can utilize deep 3D face reconstruction method\cite{Deng_Yang_Xu_Chen_Jia_Tong_2019} to extract 3DMM coefficients from a single facial image.

\subsection{Audio2Motion}
We adopted a flow-enhanced variational auto-encoder that can effectively and accurately learn the transformation from audio to motion, as proposed in \cite{ye2024realdportrait}. Considering that a 3DMM face mesh is determined by identity and expression coefficients, the identity coefficients are fixed when reconstructing the 3D face mesh of the same individual. Therefore, the reconstruction of the 3D face mesh relies solely on the expression coefficients.

\subsection{Depth map}
The identity and expression coefficients of a single facial image can be obtained using the 3DMM fitting method, or they can be directly predicted through neural networks. In this paper, we adopt the method proposed in \cite{Deng_Yang_Xu_Chen_Jia_Tong_2019} to predict identity and expression coefficients of 3DMM model from a single image. During data preprocessing, the predicted expression and identity coefficients are used to generate face 3D meshes, from which facial depth maps are obtained through differentiable rendering. During inference, the expression coefficients predicted by the A2M model trained by Real3DPortait\cite{ye2024realdportrait} replace the original expression coefficients to generate the corresponding facial depth maps, as shown in Fig. \ref{fig:framework}.

\subsection{Edit talking face}
\textbf{Encoding Images into Latent Representations} The process begins with using a pre-trained image encoder to map input images from pixel space to a lower-dimensional latent space. This dimensionality reduction significantly reduces the computational load for subsequent steps. 

\textbf{Single-step UNet in Latent Space} We adopt a single-prediction UNet architecture, similar to \cite{musetalk}. Unlike diffusion models, which strat from random noise, our approach performs predictions based on the current video frame with the mouth region occluded, aiming to generate lip movements that align with the given audio. This strategy reduces the complexity of predicting full-face information, as is typical in traditional diffusion models. Furthermore, it leverages additional reference images as input to provide reliable facial context, thereby enhancing the accuracy of facial information generation for the target subject.

Based on human intuition, the correlation between audio signals and a speaker's facial movements tends to be relatively weak, particularly for head pose changes and eye blinks. This work prioritizes the direct relationship between audio signals and lip movements. To better model the dynamics of the mouth region driven by audio, we incorporate depth information from the mouth area. This approach is designed to reinforce the alignment between audio signals and articulatory movements of the mouth.

At this stage, using the UNet single-step prediction approach, the inputs consist of the target frame with the occluded mouth region $\mathcal{I}_{tgt}^{mask}$, a randomly selected reference frame $\mathcal{I}_{ref}$ that is 
$T$ frames away from the target frame, the depth map of the lip region for the target frame $\mathcal{I}_{tgt}^{lip}$, and the audio signal $\mathcal{A}$. After encoding the image data using a VAE, the resulting features are labeled as $\mathcal{IF}_{tgt}^{mask}$, $\mathcal{IF}_{ref}$ and $\mathcal{IF}_{tgt}^{lip}$, respectively.  While the audio signal is encoded as $\mathcal{AF}$ using Whisper\cite{radford2022robustspeechrecognitionlargescale}. The three image features are then concatenated along the VAE feature channel dimension to form the input data for the UNet structure. 
\begin{equation}
\mathcal{F}_{unet} = Concat(\mathcal{IF}_{tgt}^{mask}, \mathcal{IF}_{ref}, \mathcal{IF}_{tgt}^{lip})
\label{eq:unet_input}
\end{equation}
The audio feature $\mathcal{AF}$ interacts with these image features through a cross-attention mechanism.

\textbf{Decoding Latent Representations to Images} After the single-step prediction in the latent space, the resulting latent representations are decoded back to the image space using a decoder to produce output images.

\subsection{Training Details}\label{section:train}

\textbf{Loss Function} 
Since the resolution of the latent space is relatively low and insufficient to capture fine facial details, we employ the L1 loss function in both the latent space and the image space. $\mathcal{L}_{latent}$ measures the L1 distance between the VAE-encoded features of the ground truth and the predicted frames. And $\mathcal{L}_{pixel}$ calculates the L1 distance between the two in the normalized image domain.

\begin{equation}
\mathcal{L}_{total} = \lambda_1 \mathcal{L}_{latent} + \lambda_2 \mathcal{L}_{pixel}
\label{eq:loss2}
\end{equation}

\textbf{Depth Information Selection} We adopt 3D face reconstruct method\cite{Deng_Yang_Xu_Chen_Jia_Tong_2019} to predict identity and expression coefficients from video frames. Since only lower half of faces is being predicted, the depth information is retained solely for the mouth area in the facial depth map, while the depth values for all other regions are set to zero. The mouth region in this paper is determined based on the 80 key points that delineate the mouth area, as defined by Mediapipe\cite{lugaresi2019mediapipe}. In practice, audio-driven 3D face meshes can sometimes exhibit misalignment in the mouth region relative to the real face. To improve the spatial alignment between the predicted mouth depth information and the original face image during inference, random displacement perturbations are applied to the mouth region depth information during training. This is aimed at reducing misalignment in the audio-driven lip movement generation process.  Additionally, we regard the sequence of mouth depth maps as a strong supervisory signal. Therefore, during training, depth information is randomly omitted in 50\% of the cases, to allow the model to fully leverage the audio features and learn the relationship between audio signals and mouth movements more effectively.

\section{Experiments}
\subsection{Data Preprocessing}
The training set used in the experiments consists of the open-source HDTF dataset and our curated dataset. During data preprocessing, we first segmented the videos to retain only clips containing a single face, discarding any segments with no faces or multiple faces. We then used the MTCNN detector\cite{Zhang_2016} to extract five facial key points as input to the deep 3D reconstruction model for predicting 3DMM coefficients. Finally, we use DWPose\cite{Yang_Zeng_Yuan_Li_2023} to extract facial bounding boxes, which were used to crop the faces and depth maps. For longer videos, frames are extracted at 25 frames per second, and 10,000 frames are randomly selected to ensure the model's generalization ability.

\subsection{Experimental Setups}
\textbf{Implementation Details.} Our models are trained from scratch using 8 NVIDIA H800 GPUs for one day. All input images are resized to a resolution of 256x256. The batch size is set to 128. Additionally, the gradient accumulation step is set to 1. The Adam optimizer is employed with a learning rate of $1 \times 10 ^ {-5}$.  The loss weights for the latent feature space and the pixel space are set to $\lambda_1=2$ and $\lambda_2=1$, respectively.

\textbf{Evaluation Metric.} Given the absence of ground-truth talking face videos, we employed FID\cite{heusel2018ganstrainedtimescaleupdate} to assess the visual quality of generated videos. To evaluate the synchronization between lip movements and audio, we used LSE-C and LSE-D proposed in Wav2Lip\cite{Prajwal_Mukhopadhyay_Namboodiri_Jawahar_2020}. Following the unpaired evaluation settings described in Wav2Lip, we select a video clip and an audio clip from different videos to perform synthesis. For the HDTF dataset and our collected dataset, we generate approximately 500 and 900 audio-video pairs, respectively, with each audio-video pair lasting about 10 seconds.

\textbf{Baseline.} In our quantitative comparative experiments, we selected several open-source implementations for comparison, including Wav2Lip\cite{Prajwal_Mukhopadhyay_Namboodiri_Jawahar_2020} and MuseTalk\cite{musetalk}. We modified the dataset code of Wav2Lip\footnote{https://github.com/Rudrabha/Wav2Lip}, first retraining the lip-sync expert model and then training the generation model. To achieve better results, we referred to MuseTalk\footnote{https://github.com/TMElyralab/MuseTalk} and implemented the training code to generate talking face videos. Comparative experiments were conducted on the two datasets mentioned above.

\subsection{Quantitative Results}
\begin{figure*}
    \centering
    \begin{subfigure}{0.45\textwidth}   % 设置子图的宽度
        \centering
        \includegraphics[width=\linewidth]{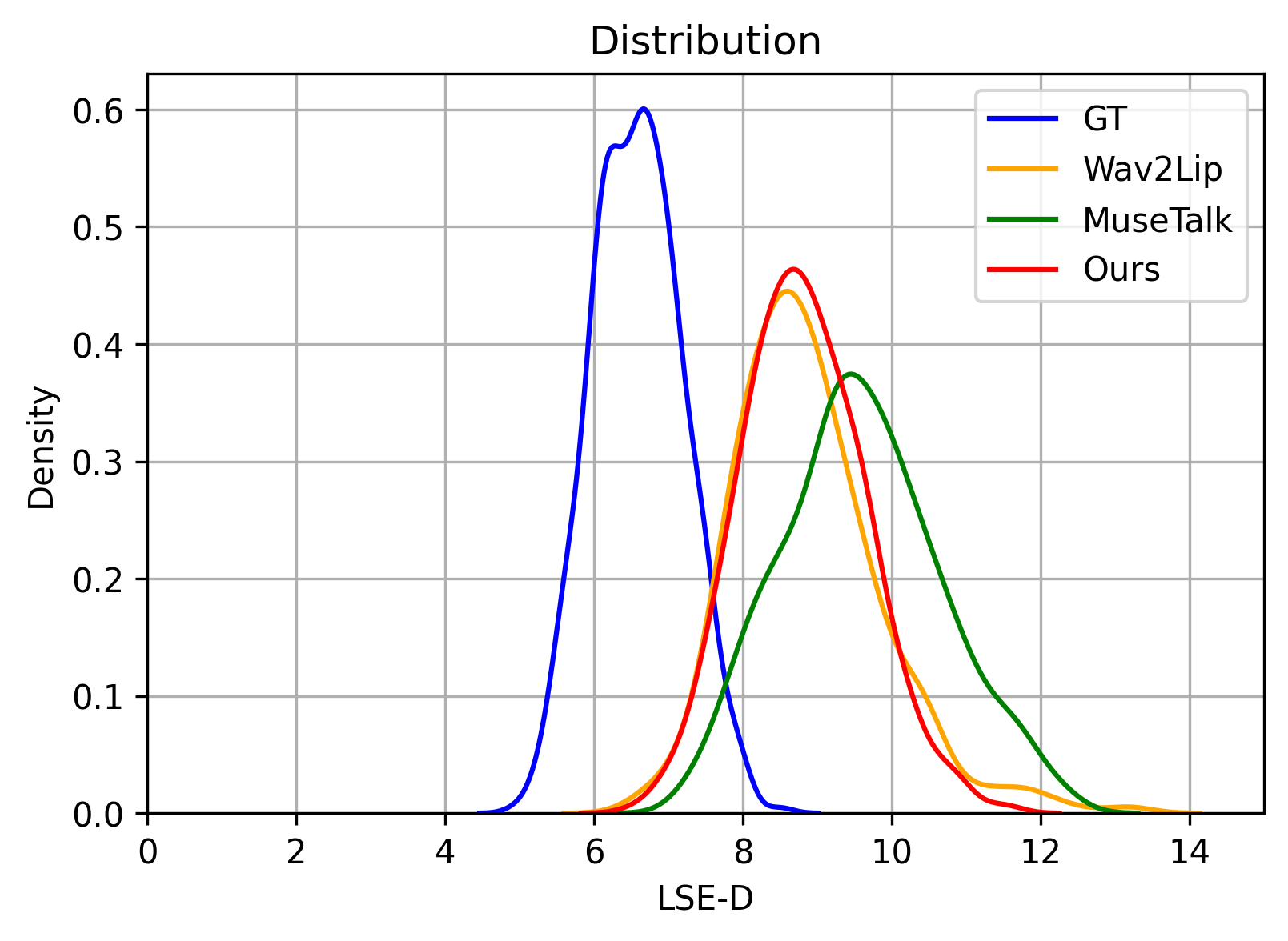}  % 替换为图片路径
        \label{fig:hdtf_lse_dist}
    \end{subfigure}
    \hfill
    \begin{subfigure}{0.45\textwidth}
        \centering
        \includegraphics[width=\linewidth]{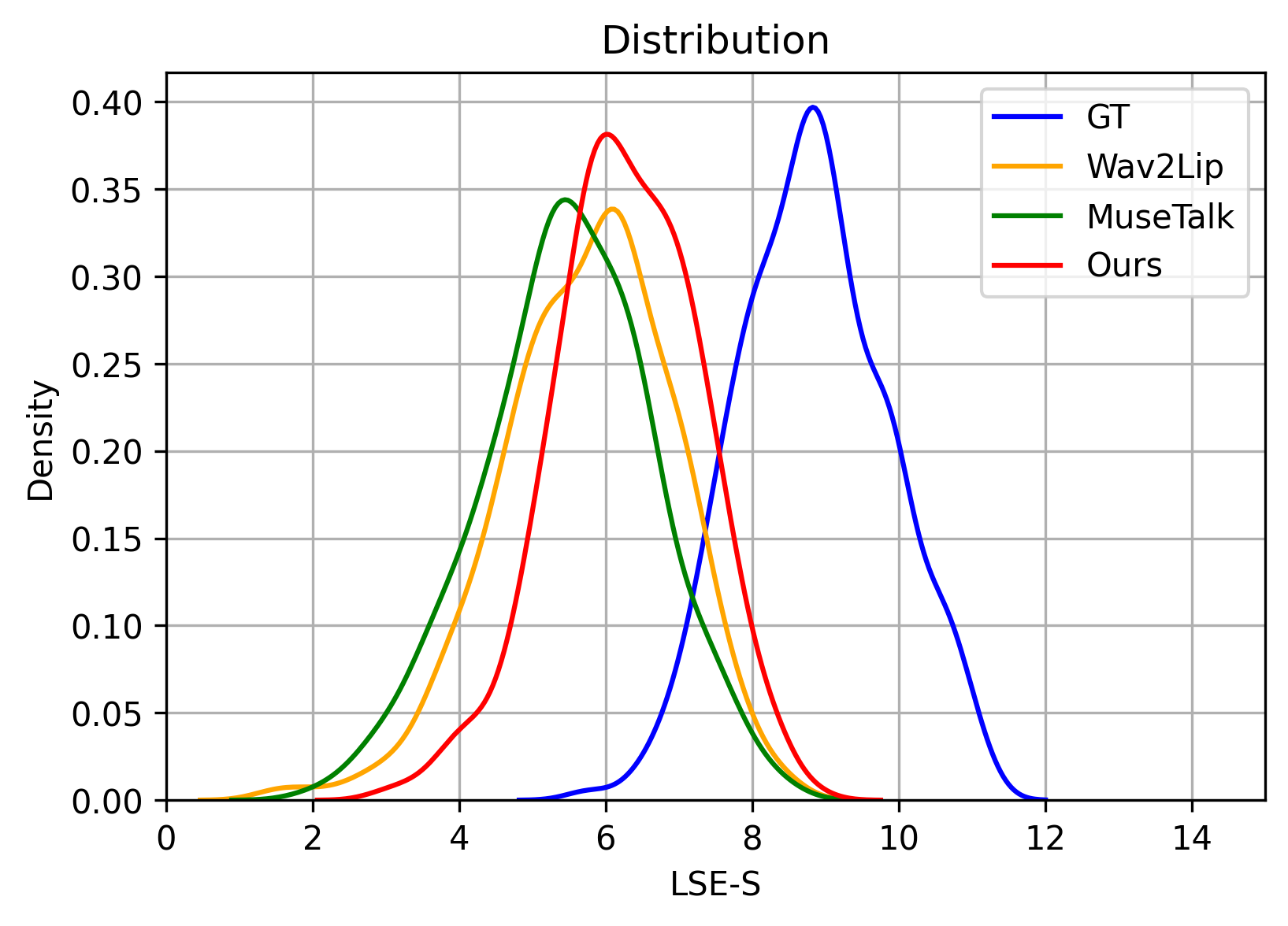}  % 替换为图片路径
        \label{fig:hdtf_lse_sim}
    \end{subfigure}
    \caption{The distribution curves of LSE-D and LSE-C scores on the HDTF dataset}
    \label{fig:d1}
\end{figure*}

\textbf{Comparison on the HDTF dataset.} Table \ref{tab:metrics_comparison_hdtf} presents the quantitative evaluation results on the HDTF dataset, showing that our proposed DeepTalkFace method outperforms other methods across all metrics. Additionally, the distribution curves of LSE-D and LSE-C scores for each method, displayed in Fig. \ref{fig:d1}, provide further evidence of DeepTalkFace's superior performance.

\begin{table}[htbp]
\centering
\caption{The quantitative comparisons on the HDTF dataset.}
\label{tab:metrics_comparison_hdtf} 
\begin{tabular}{lllll} 
\toprule
\textbf{Method} & {\textbf{FID$\downarrow$}} & {\textbf{LSE-D$\downarrow$}} & {\textbf{LSE-C$\uparrow$}} \\
\midrule
Wav2Lip & 38.98 & 8.86 & 5.74 \\
MuseTalk & 6.87 & 9.60 & 5.47 \\
Ours & \textbf{6.76} & \textbf{8.81} & \textbf{6.27} \\
\bottomrule
GT & / & 6.58 & 8.83 \\
\bottomrule
\end{tabular}
\end{table}

\begin{figure*}
    \centering
    \begin{subfigure}{0.49\textwidth}   % 设置子图的宽度
        \centering
        \includegraphics[width=\linewidth]{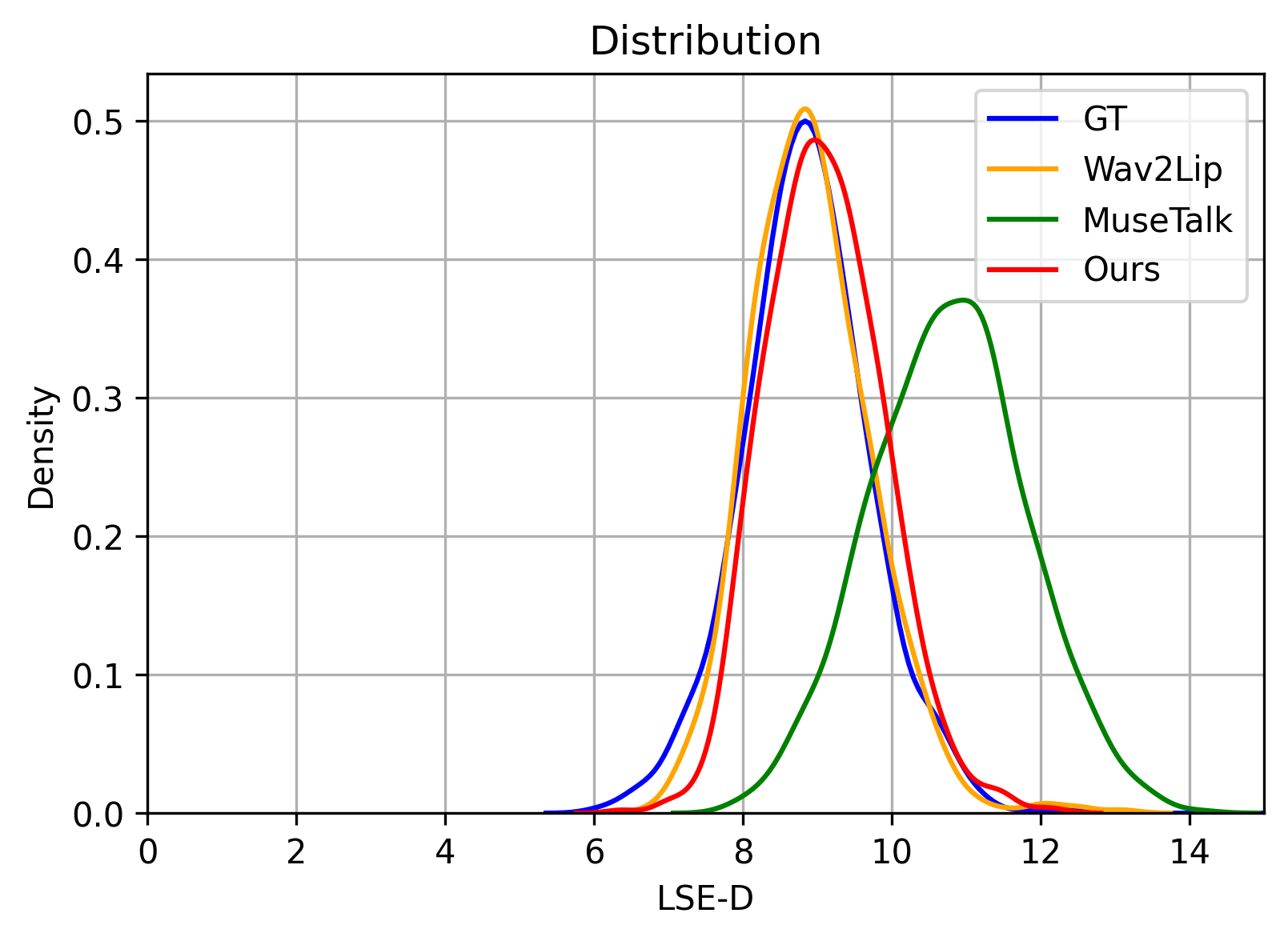}  % 替换为图片路径
        \label{fig:dybili_lse_dist}
    \end{subfigure}
    \hfill
    \begin{subfigure}{0.49\textwidth}
        \centering
        \includegraphics[width=\linewidth]{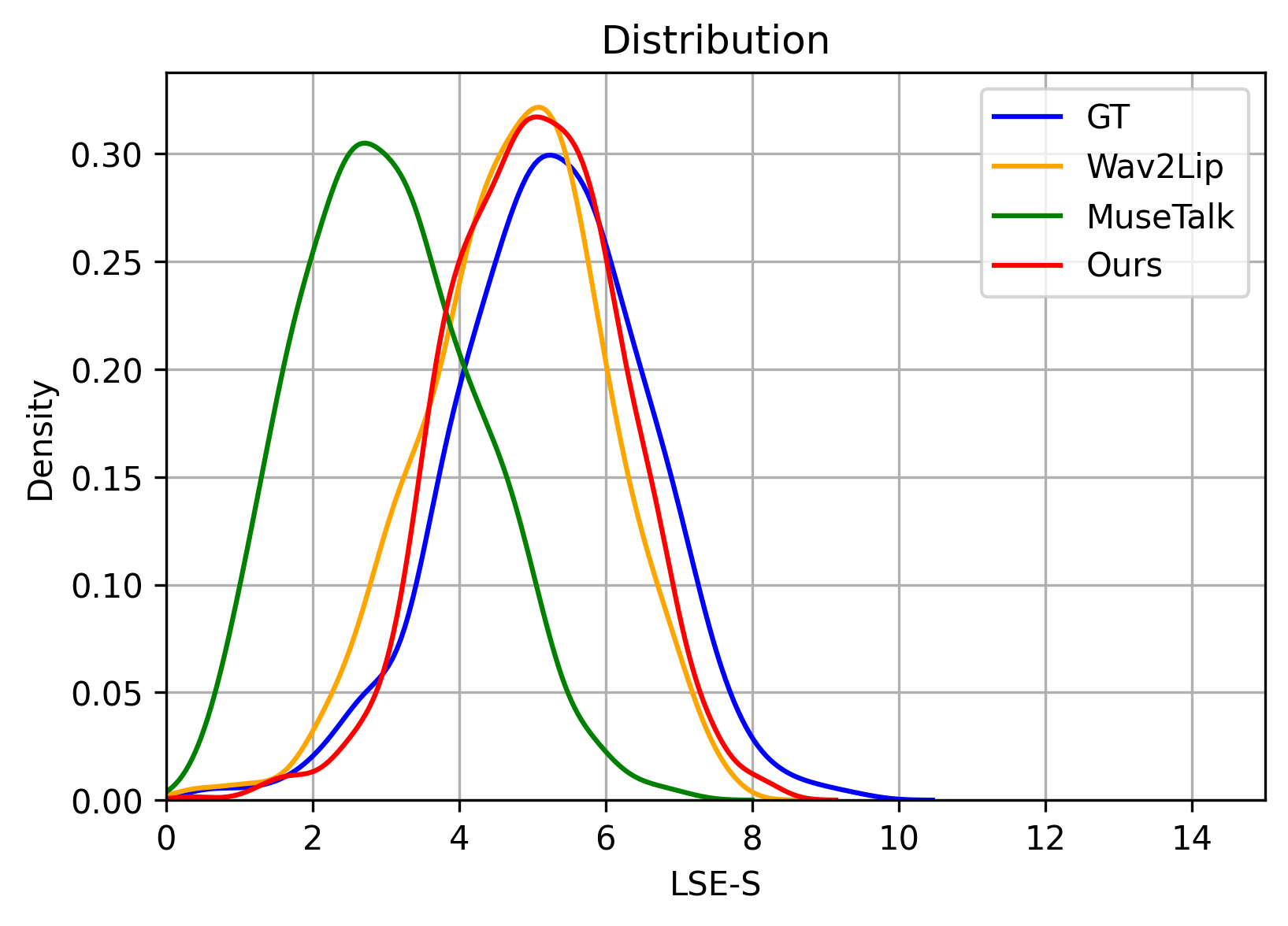}  % 替换为图片路径
        \label{fig:dybili_lse_sim}
    \end{subfigure}
    \caption{The distribution curves of LSE-D and LSE-C scores on our collected dataset}
    \label{fig:d2}
\end{figure*}

\textbf{Comparison on our collected dataset.} We further evaluate the proposed JoyGen and other methods on our collected datasets. As shown in Table \ref{tab:metrics_comparison_collected_dataset}, JoyGen achieves the lowest FID score at 3.19, indicating a significant improvement in visual quality compared to existing methods. Additionally, our method exhibits strong lip-audio synchronization, with LSE-D and LSE-C scores closely matching with ground truth values on this dataset. The distribution curves of LSE-D and LSE-C scores for each method, illustrated in Fig. \ref{fig:d2}, further the superior synchronization performance achieved by JoyGen.

\begin{table}[htbp]
\centering
\caption{The quantitative comparisons on our collected dataset.}
\label{tab:metrics_comparison_collected_dataset} 
\begin{tabular}{lllll} 
\toprule
\textbf{Method} & {\textbf{FID$\downarrow$}} & {\textbf{LSE-D$\downarrow$}} & {\textbf{LSE-C$\uparrow$}} \\
\midrule
Wav2Lip & 7.51 & \textbf{8.91} & 4.37 \\
MuseTalk & 3.36 & 10.76 & 3.04 \\
Ours & \textbf{3.19} & 9.11 & \textbf{5.02} \\
\bottomrule
GT & / & 8.87 & 5.24 \\
\bottomrule
\end{tabular}
\end{table}

\subsection{Qualitative Results}
\begin{figure*}
    \centering
    \begin{subfigure}{0.48\textwidth}   % 设置子图的宽度
        \centering
        \includegraphics[width=\linewidth]{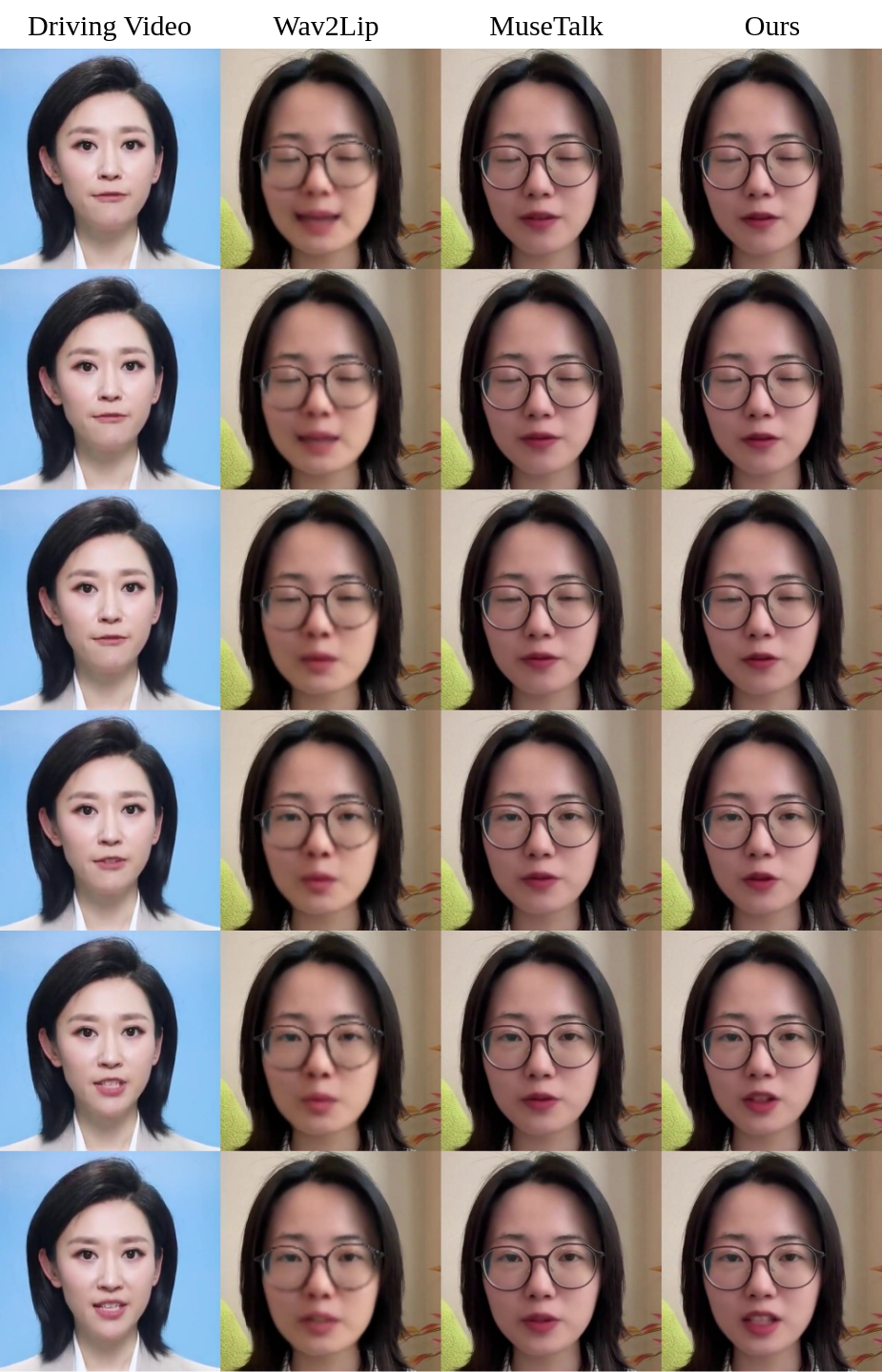}  % 替换为图片路径
        \label{fig:qualitative_evaluation}
    \end{subfigure}
    \hfill
    \begin{subfigure}{0.48\textwidth}
        \centering
        \includegraphics[width=\linewidth]{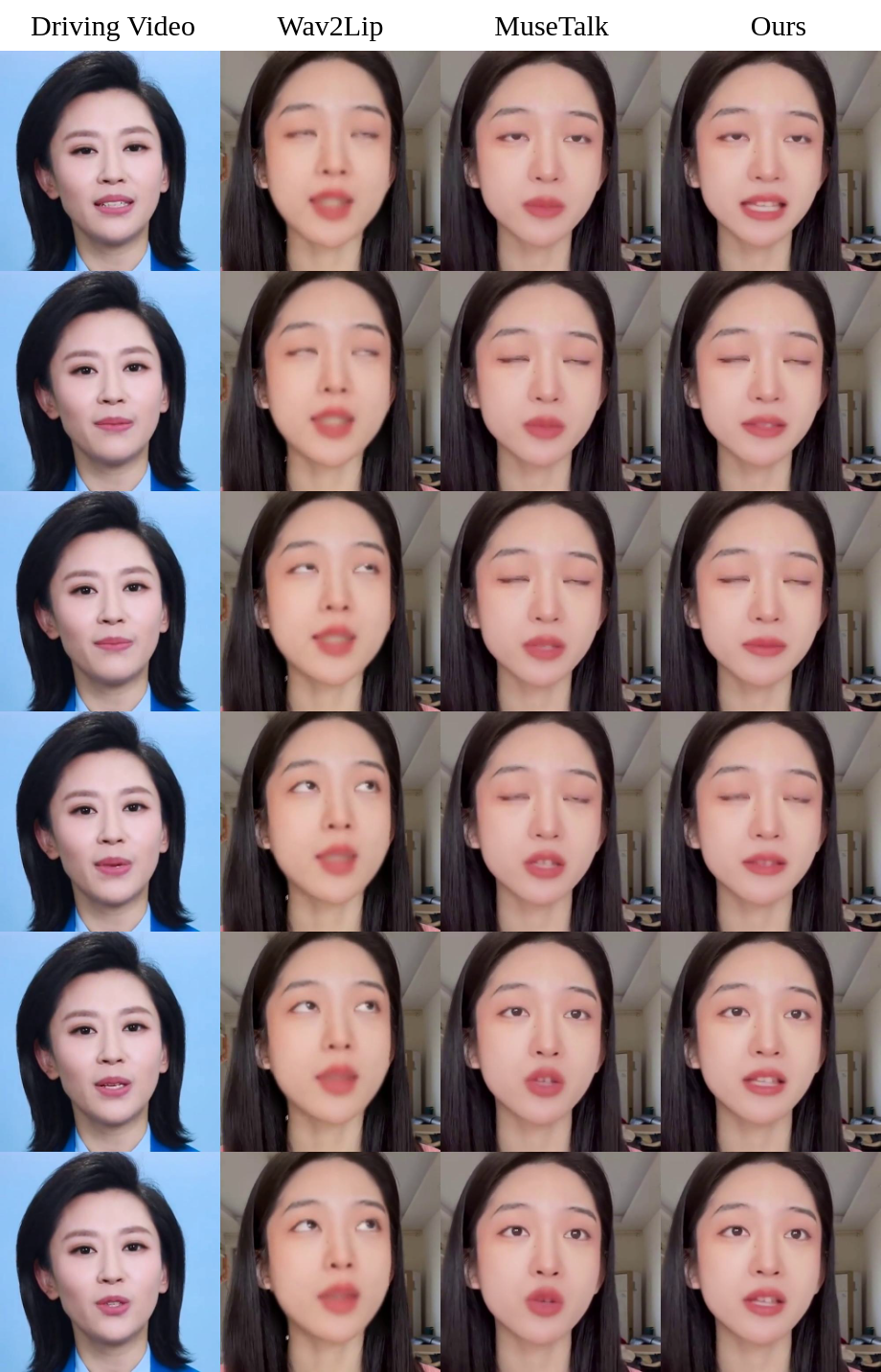}  % 替换为图片路径
        \label{fig:qualitative_evaluation2}
    \end{subfigure}
    \caption{Qualitative comparisons across different methods show that our approach significantly outperforms others in terms of lip synchronization and image quality.}
    \label{fig:qualitative}
\end{figure*}

To evaluate lip shape synchronization and visual quality more intuitively, we conducted a comparative analysis of several methods. Fig. \ref{fig:qualitative} illustrates the lip dynamics in each generated video alongside the reference driving video. Each column contains six consecutive frames from the videos. Our method achieves higher video quality and more precise lip movements. In the left driving video, the lip transitions from open to closed and back to open; in the right, it goes from closed to open. Only our method maintains consistent lip shape alignment with the driving video. Although Wav2Lip-based videos have better LSE-D and LSE-C scores, Wav2Lip tends to generate blurred-out mouth region.

\section{Conclusions and Future Work}
In this paper, we introduce a method to achieve lip-audio synchronization by leveraging both face depth maps and audio features. To achieve more precise control over lip shape editing in the video and to avoid introducing additional noise, we employ a single-step UNet for generating the facial region. As demonstrated above, our method achieves higher video quality and more precise lip-audio synchronization.

Compared to diffusion models, which generate high-resolution facial images with rich details through multiple denoising steps, single-step UNet predictions still have room for improvement in terms of image clarity and detail. Incorporating additional mouth depth information can further enhance the synchronization between lip movements and audio. Similarly, integrating explicit eye-blink signals enables precise control over blinking actions in the original video, enriching the synthesized content with natural eye expressions. In the future, to enhance the visual quality of synthesized videos and improve audio-lip synchronization, we plan to adopt multi-step UNet predictions or employ more advanced network architectures. Additionally, we aim to train a more robust model for predicting lip movements from audio signals.

%%%%%%%%% REFERENCES
{\small
\bibliographystyle{ieee_fullname}
\bibliography{egbib}
}
% \end{multicols}
\end{document}